\documentclass[twoside,11pt]{article}

\usepackage{jmlr2e}
\usepackage{xcolor}
\usepackage{indentfirst}
\usepackage{authblk}

\ShortHeadings{Convolutional Neural Networks for Medical Diagnosis from Admission Notes}{ }
\firstpageno{1}

\begin{document}

\title{Convolutional Neural Networks for Medical Diagnosis from Admission Notes} 


\author[1]{Christy Yuan Li}
\author[1]{Dimitris Konomis}
\author[1]{Graham Neubig}
\author[1]{Pengtao Xie}
\author[1,2]{Carol Cheng}
\author[1,3]{Eric Xing}
\affil[1]{\small{Petuum Inc., Pittsburgh, 15222, USA}}
\affil[2]{\small{Department of Psychiatry, University of Pittsburgh Medical Center, Pittsburgh, 15213, USA}}
\affil[3]{\small{eric.xing@petuum.com}}

\maketitle
  
\begin{abstract} 
  
\textbf{Objective}
Develop an automatic diagnostic system which only uses textual admission information from Electronic Health Records (EHRs) and assist clinicians with a timely and statistically proved decision tool. The hope is that the tool can be used to reduce mis-diagnosis.   

\textbf{Materials and Methods}
We use the real-world clinical notes from MIMIC-III, a freely available dataset consistsing of clinical data of more than forty thousand patients who stayed in intensive care units of the Beth Israel Deaconess Medical Center between 2001 and 2012 ~\citep{johnson2016mimic}. We proposed a Convolutional Neural Network model to learn semantic features from unstructured textual input and automatically predict primary discharge diagnosis. 

\textbf{Results}
The proposed model achieved an overall 96.11\% accuracy and 80.48\% weighted F1 score values on 10 most frequent disease classes, significantly outperforming four strong baseline models by at least 12.7\% in weighted F1 score.

\textbf{Discussion} 
Experimental results imply that the CNN model is suitable for supporting diagnosis decision making in the presence of complex, noisy and unstructured clinical data while at the same time using fewer layers and parameters that other traditional Deep Network models.

\textbf{Conclusion}
Our model demonstrated capability of representing complex medical meaningful features from unstructured clinical notes and prediction power for commonly misdiagnosed frequent diseases. It can use easily adopted in clinical setting to provide timely and statistically proved decision support.

\end{abstract}

\section{Background and Significance}

Mis-diagnosis is one of the most severe problems in healthcare~\citep{diagnostic_error}, inducing significant harm to patients' well-being. As reported by~\citep{12million}, approximately 12 million adults are misdiagnosed in outpatient medical care, which amounts to 1 out of 20 adult patients. More importantly, around 40,000 - 80,000 patients die annually in the U.S.A due to diagnostic errors, as evidence from autopsies~\citep{diagnostic_error} indicates.

While a plethora of factors can lead to mis-diagnosis, untimely diagnosis and lack of guidelines stand out as the most significant factors \citep{misdiagnosis}. Many diseases such as asthma, chronic obstructive pulmonary disease and bronchiectasis, share similar symptoms such as dyspnea, coughing, wheezing and expectoration, with both factors rendering their differential diagnosis a difficult task. However, differentiating between these diseases in the early stage, such as at patient's admission time, is extremely important since different treatments could lead to very different clinical outcomes, and the adoption of an improper treatment plan could prove to be disastrous for the patient's health. Additionally, insufficiency of clinical guidelines for handling rare symptoms quite often can result in failure to correctly explain a patient's symptoms.  As a result, no suitable treatments are immediately administered to the patient. 

The significant improvement of Electronic Health Record (EHR) systems over the past decades has facilitated standardization of data collection by health-care professionals and allowed clinicians to access patients' important and  clinical information at their earliest convenience, preventing mis-diagnosis or delayed diagnosis. However, initial diagnostic decision-making based on EHRs still faces some challenges. The absence of a protocol for filling an EHR as well as the fact that an EHR usually contains information collected from multiple clinicians, with different personalized writing styles who potentially work at different health institutions, results in EHRs not sharing the same length or writing style, missing structure, alternating between official names, unofficial names, medical names and abbreviations of diseases and containing various misspellings and grammar errors. According to a recent study ~\citep{sinsky2016allocation}, clinicians were found spending up to half of their total working time on EHR and desk work and less than a third of their time interacting directly with patients. 

Convolutional Neural Network (CNN) models' ability to identify patterns and learn appropriate latent representations from  textual data has rendered them as one of the most powerful models for classification. 
In this work, we investigated the potential of a CNN model to support the clinician's diagnostic decision making, formulating discharge diagnosis prediction as a multiclass classification problem. Our model takes as input only a subset of the information contained in a patients EHR upon admission and produces a discharge diagnosis prediction in its output. We trained a separate neural network in order to learn the \textit{embeddings} (low dimensional real valued vectors that act as features) of the words in the vocabulary of our dataset and used the weights of this model to initialize the weights of the embedding layer of the CNN model. This is a critical optimization that further boosted the performance of the CNN model.

MIMIC-III is a freely available dataset containing clinical data of more than forty thousand patients who stayed in intensive care units of the Beth Israel Deaconess Medical Center between 2001 and 2012 ~\citep{johnson2016mimic}. After appropriate preprocessing, which included the application of the coreference resolution method on the disease names (seeks to find the mentions in text that refer to the same real-world entity), extracting the clinical notes that correspond to the most frequent diseases (and ignoring the rest), discharge diagnosis prediction was formulated as a multiclass classification problem, with the following 10 classes: coronary artery disease, hemorrhage, pneumonia, myocardial infarction, gastrointestinal bleeding, fracture, aortic stenosis, cardiac failure, prematurity and stroke. We evaluated the performance of our CNN model as well as other baseline classification models such as Support Vector Machines (SVM), Random Forest (RF), Multi Layer Perceptron (MLP) and Logistic Regression (LR) on the MIMIC-III dataset using precision, accuracy, recall and F1 score metrics for each of the 10 different diseases. 
Experiments indicate that our CNN model, achieving overall 96.11\% accuracy and 80.84\% weighted F1 score, outperforms all of the baseline models with respect to 9 out of the 10 disease classes. Furthermore the F1 score value of 80.84\%, at least 12\% higher than that of the best of the baseline models, reflects that the CNN model performed equally well on all disease classes independently of how frequently they appear and are diagnosed in the MIMIC corpus. We hence believe it be a great tool for discharge diagnosis support.

\section{Methods}

\subsection{CNN-based multi-class text classification } 

\begin{figure}
  \includegraphics[width=\linewidth]{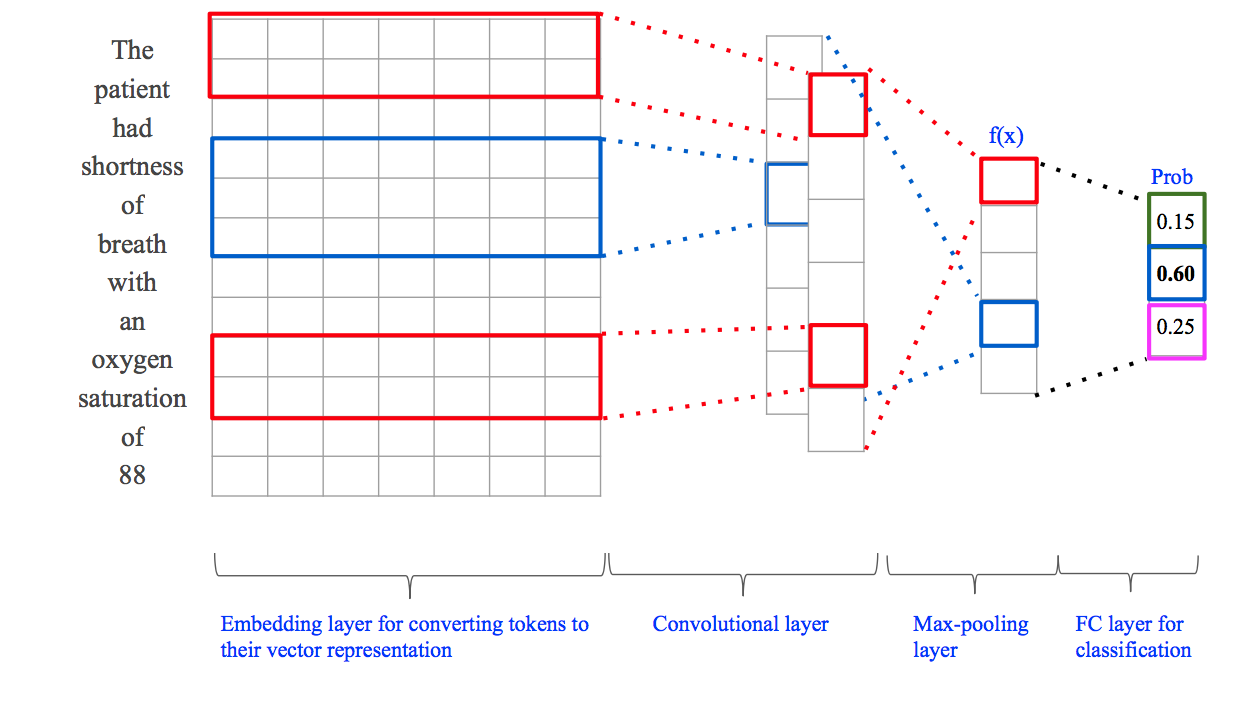}
  \caption{Demonstration of CNN model which consists of four components: embedding layer, convolutional layer, max-pooling layer, and fully-connected layer. }
  \label{fig:cnn-model}
\end{figure}

The discharge diagnosis prediction problem is cast as a multi-class text classification problem, with an admission note (after appropriate preprocessing) being fed into the input of a convolutional neural network, and classified by the latter into one of $K$ primary discharge diseases. We now shed light on the different parts of the models' architecture, as depicted in figure \ref{fig:cnn-model}.

The CNN among other things, learns an underlying representation of an admission note in a high-dimensional space. The architecture of the CNN model is depicted in Figure~\ref{fig:cnn-model}. The embedding layer maps each word in the admission note to its embedding, (a low-dimensional real-valued vector) and acts essentially as a lookup table. The embedding layers can be represented by a $V \times E$ matrix, where $V$ is the number of different words of the vocabulary considered and $E$ the dimension of an embedding. The embedding layer outputs a $L \times E$ matrix for an admission note consisting of $L$ words.

The \textit{convolutional} layer consists of $F$ independent filtering operations, ($F$ is usually 128 or 256) with each filter trying to extract different type of information from the $L \times E$ embedding representation of the admission note. The $i$-th filtering operation can be thought of as sliding a $H_i \times E$ matrix (filter) through the output of the embedding layer and computing its dot product with the corresponding $H_i \times E$ area of the admission note's embedding representation. Performing this operation trivially, the $i$-th filtering operation will output an $(L-H_i+1)$ dimensional vector; padding the output of the embedding layer by $H_i-1$ rows of zeros, we ensure the $i-$th filtering operation's output is an $L$ dimensional vector. The output of the convolutional layer results from stacking together these vectors to obtain a $L \times F$ matrix; this is the ``feature representation" that the CNN is learning.

The output of the convolutional layer, an $L \times F$ matrix, is then fed into a \textit{max-pooling layer}. The $i$-th max-pooling operation selects the maximum value of the $i$-th convolution ($i$-th column of the matrix). The output of the max-pooling layer is thus an $F$-dimensional vector, having a max-value per filter. A CNN usually contains filters of different sizes that capture patterns across contexts of different sizes (number of words). Our model uses an embedding of $E=128$ and total of $F_1=64$ filters of size $H_1 \times E = 3 \times 128$, $F_2=64$ filters of size $H_2 \times E = 4 \times 128$ and $F_3=64$ filters of size $H_3 \times E = 5 \times 128$.

The $F$-dimensional output of the max-pooling layer enters a \textit{fully-connected} layer. This is essentially a layer of $F$ input neurons connected to all $K$ output neurons. Complex co-adaptations of the fully-connected layer's weights on the training data can cause the CNN model to overfit. To prevent overfitting, individual neurons in the fully-connected layer are either kept with probability $p$ or ``dropped out" with probability $(1-p)$, a technique known as dropout.

A final \textit{softmax} layer converts the $K$-dimensional output of the fully-connected layer, $x$ into a $K$-dimensional probability distribution vector $\pi$, applying essentially the normalized exponential function to $x$:

\begin{equation}
  \pi_j = \frac{e^{x_j}}{\sum_{k=1}^{K}e^{x_k}}, k=1,2, \dots K
\end{equation}

Finally, the true class of a clinical note is represented as a ``one-hot" $K$-dimensional vector $y$, with $y_i=1$ if $i$ corresponds to the disease that was actually diagnosed and $y_i=0$ otherwise. The CNN is then trained, end-to-end, using the stochastic gradient descent algorithm in order to minimize the \textit{cross-entropy} loss of $y$ and $\pi$:

\begin{equation}
  CE(y, \pi) = -\sum_{k=1}^{K} y_k \log \pi_k
\end{equation}

\subsection{Word embedding pre-training} 
We employed Skip-gram model~\citep{mikolov2013distributed} for training medical word embeddings for initializing the CNN-based text classification model. This is beneficial because the semantic information of clinical notes can be incorporated through the pre-trained embeddings of words from clinical notes. We trained the embedding model on the whole MIMIC III dataset where all data fields including admission information and non-admission information sush as brief hospital course, discharge instructions, discharge plan, discharge medication, lab test during hospitalization are used. The Skip-gram model is designed to learn the probability distribution of words that appear closely in the corpus. The objective function is shown as follows: 

\begin{equation}
\frac{1}{T} \sum_{t=1}{T} \sum_{-c \le j \le c, j \ne 0} log P (\textbf{w}_{t+j} \| \textbf{w}_t) 
\end{equation}

where the $T$ is the number of words in the input sequence, $c$ is the size of the training context. The conditional probability of a target word given current word is defined by softmax function: 

\begin{equation}
P (\textbf{w}_{t+j} \| \textbf{w}_t) = \frac{exp(\textbf{v}_{wO}^{T} \textbf{v}_{wI})}{\sum_{w=1}^{W} exp(\textbf{v}_{wO}^{T} \textbf{v}_{wI})}
\end{equation}

where the $W$ is the vocabulary size, and $v_{wI}$ and $v_{wO}$ are the input and output vector representations of word w. 

In practice, we used fasttext ~\citep{joulin2016fasttext} to train the embedding model. Wrods whose frequency is less than or equal to 1 were removed. During initializing of the classification model, we used embedding vectors of in-vocabulary words, and computed the vector representations of out-vocabulary words using the function provided by the Fasttext. 
 
\subsection{Disease coreference resolution} 

\begin{figure}
\centering
  \includegraphics[width=100mm]{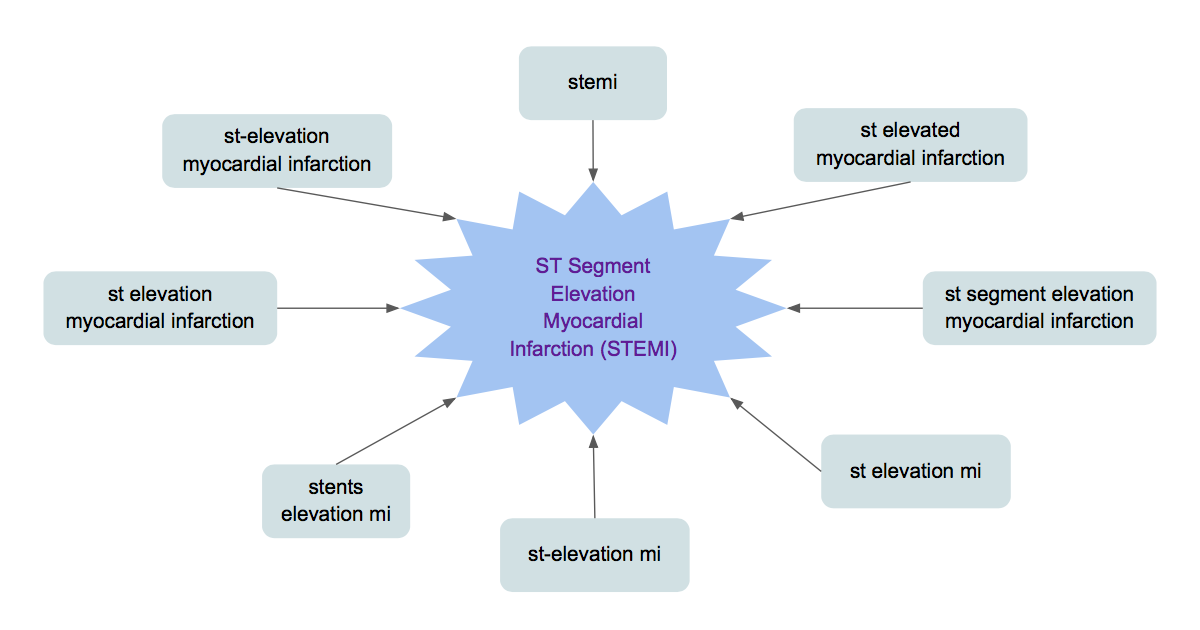}
  \caption{An example of disease coreference resolution of ST Segment Elevation Myocardial Infarction}
  \label{fig:disease-coreference-example}
\end{figure}

Real-world clinical notes are characterized by lack of structure and the usage of different words referring to the same term. This is a result of doctors' personalized writing styles as well as their tendency to alternate between official names, abbreviations, unofficial names and medical custom names when they refer to a particular term. Moreover, various misspellings and inconsistency in using lowercase/uppercase letters further increase the variety of different words used to refer to the same term.
Figure \ref{fig:disease-coreference-example} illustrates an example where the terms ``segment elevation myocardial infarction", ``stents elevation myocardial infarction", ``st-elevation myocardial infarction",   ``st elevation myocardial infarction", ``st elevated myocardial infarction", ``st segment elevation myocardial infarction", ``st elevation mi", ``st-elevation mi" and ``stemi" all refer to the disease with official name ``ST Segment Elevation Myocardial Infarction". 

\textit{Coreference resolution}, the task of finding all expressions that refer to the same entity in a text, is an important step for a lot of higher level NLP tasks that involve natural language understanding such as document summarization, question answering, and information extraction. Although there exists a plethora of machine learning algorithms for coreference resolution, both supervised and un-supervised, we used a remarkably simple, fast and still successful approach on the MIMIC dataset: we collected all discharge diagnosis disease names appearing after phrases such as ``diagnosis", ``primary diagnosis" and: (i) manually grouped words that refer to the same disease, (ii) replaced all the words appearing in the groups by the name of the disease.

The MIMIC III dataset includes the International Statistical Classification of Diseases and Related Health Problems (ICD-9) codes associated with the disease(s) of each clinical note. It is important to note that even though ICD-9 is very fine-grained and contains the official names of diseases and as many sub-diseases as possible, the clinical records reflect the doctors' tendency to not closely follow ICD-9 terminology. In a preliminary experiment, we measured the classification performance of our CNN model under two different settings with respect to the classification labels used: in the first setting, we used the disease names that result from coreference resolution whereas in the second setting ICD-9 official disease names. Given that the classification accuracy was about 5\% higher in the first setting, we chose the disease names that resulted from our simple coreference resolution method as the classes for the multi-class classification problem.

\section{Experiments \& Results}

\subsection{Data} 
\subsubsection{Dataset}

We evaluated the performance of the proposed CNN model using MIMIC-III, a freely available dataset that consists of clinical data for more than 40,000 patients who stayed in the intensive care units of the Beth Israel Deaconess Medical Center between 2001 and 2012~~\citep{johnson2016mimic}. The dataset consists of more than 50,000 anonymized clinical records that include, among others, demographics, chief complaints, past medical history, vital signs, procedures, lab tests, medications and discharge diagnosis information.
Given that our model tries to provide an accurate prediction of the discharge diagnosis by focusing solely on information available upon admission, we filtered information such as chief complaints, past medical history, past surgical history, social history, family history, allergies, laboratory examination results upon admission and medications taken upon admission from the clinical records.

\subsubsection{Data preprocessing}

The filtered information, initially in text format and containing abbreviations, misspellings and grammar errors, needed to undergo through several pre-processing tasks before it could be feeded into the input of the CNN model. These preprocessing tasks included the concatenation of separate lines, the removal of duplicate spaces, the splitting of words by punctuation, the transformation of words to lowercase as well as the replacement of specific numbers, person names, hospital names, dates and times by ``***" for the sake of anonymization. Finally, since the length (in number of words) was not unique among the different clinical notes that resulted from the previous preprocessing tasks and the CNN model only operates on documents of a fixed common length, an additional preprocessing task was carried out. 
This last preprocessing task involved the computation of a maximum document length as the maximum length (in number of words) not exceeded by 90\% of the clinical notes that resulted from the previous preprocessing tasks, and a truncation at exactly this length for those preprocessed clinical notes whose length was higher than this threshold value.

\subsubsection{Classification categories}

We applied coreference resolution on the MIMIC III dataset, resulting with 479 unique disease categories. We further analyzed the dataset and discovered the top 10 most frequent diseases, which we used as the labels for our classification problem. In the case of a clinical note that mentions multiple diseases, only the first one is counted, based on the assumption that the first disease is the most important. Furthermore, we discarded clinical notes which did not mention at all any of these 10 diseases, ending up with a total of 13152 EHRs. The sample distribution of the 10 diseases is illustrated in figure~\ref{tab:sample-distribution} and table ~\ref{tab:sample-distribution}.

\begin{table}
\centering
    \begin{tabular}{ |l | l| l| l|}
    \hline
    disease & number of samples\\ \hline
    Coronary artery disease & 3193 \\
    Hemorrhage & 1955 \\
    Pneumonia & 1634 \\
    Myocardial infarction & 1229 \\
    Gastrointestinal bleeding & 1158 \\
    Fracture & 1047 \\
    Aortic stenosis & 934 \\
    Cardiac failure & 927 \\
    Prematurity & 559 \\
    Stroke & 504 \\
    \hline
    \end{tabular}
    \caption{Sample distribution of the 10 disease categories. }
    \label{tab:sample-distribution}
\end{table}

\begin{figure}
\centering 
  \includegraphics[clip, trim=0.5cm 16cm 1cm 2cm, width=0.90\textwidth]{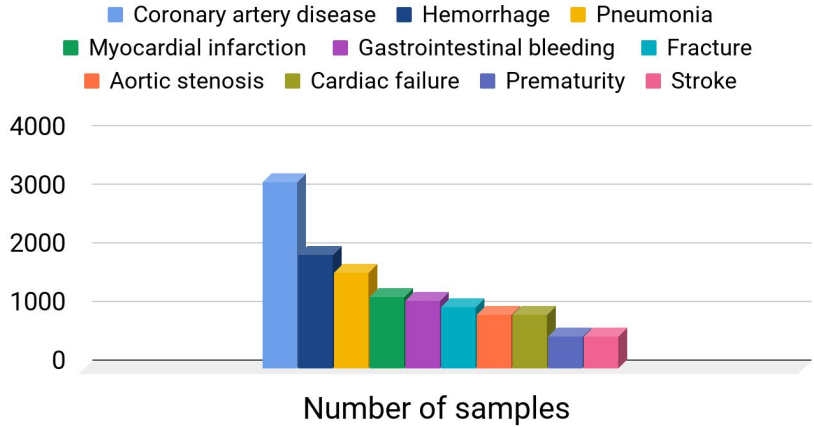}
  \caption{Bar chart of sample distribution of the 10 disease categories. We can see that the sample distribution is unbalanced. More than 23\% of samples drop in the first category coronary artery disease while less than 4\% of samples are in the last two disease categories which are prematurity and stroke.}
  \label{fig:sample-distribution}
\end{figure}

\subsection{Evaluation Approach}

For each individual disease category, we evaluated the model performance on the testing set keeping track of the following metrics: accuracy, true negative rate, false positive rate, false negative rate, precision, recall and F1 score. We additionally measured, for each of these metrics, a un-weighted and a weighted average value across the different disease categories (where the weight $w_i$ for disease category $i$ is simply the ratio of the clinical notes where disease $i$ was diagnosed over the total number of clinical notes in the testing set. The weighted average values capture well the distribution of diagnosed diseases in the testing set.

\subsection{Baseline Models}

For multi-class disease classification, we compare our CNN text classification model against several baseline models: support vector machine (SVM), random forest (RF), multi-layer perception (MLP) and logistic regression (LR). Tf-idf (term frequency-inverse document frequency) ~\citep{salton1975vector} is a common numerical statistic reflecting the importance of a word to a document in a corpus. For each clinical note, we first create a $V$-dimensional vector ($V$ being the size of our vocabulary) containing $L$ tf-idf values for the words that appear in it and $(V-L)$ zero values for the words that are absent. Therefore, for a set with $T$ clinical notes we obtain a $T \times V$ feature matrix. Given the large value of $V$, we further apply the Principal Components Analysis (PCA) dimensionality reduction technique to obtain a $T \times V'$ feature matrix with $V' \ll V$. The rows of this matrix are the actual clinical notes' feature vectors used by each of the baseline classification models during training and testing.

\subsection{Implementation Details} 
We randomly split the MIMIC III dataset into training, validation and testing sets by a ratio of 7:1.5: 1.5. The best trained model is selected based on the performance on the validation set and tested on the testing set. For each model, we further trained 5 different times with each using a different random seed for splitting the dataset, and ended up with 5 evaluation results. The 5 evaluations results were averaged to represent the final evaluation results for the corresponding model. 
  
For our CNN model, we trained the Fasttext ~\citep{joulin2016fasttext} word embedding model (using embedding dimension $E=128$) on the full MIMIC dataset and obtained an initialization of the embedding layer weights. We used a learning rate $\alpha=0.0001$, $F_1=64$ filters of size $H_1 \times E=3 \times 100$, $F_2=64$ filters of size $H_2 \times E=4 \times 128$ and $F_3$ filters of size $H_3 \times E=5 \times 100$, dropout probability $p=0.5$, the adam optimizer for gradient descent and trained on a single GPU. We used two layers with 100 and 10 neurons in each layer for MLP model, along with Relu activation functions and early stopping. We used a kernelized one-vs-all SVM with an RBF kernel and tolerance $\epsilon=0.001$.

\subsection{Results on multi-class diagnosis classification}
\begin{figure}
\centering 
  \includegraphics[clip, trim=2.9cm 18cm 3cm 3.6cm, width=0.80\textwidth]{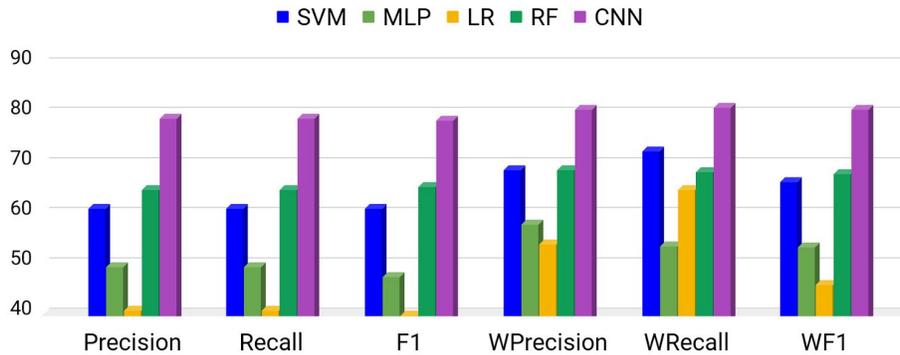}
  \caption{Bar chart demonstrating the performance difference among SVM, random forest, MLP, logistic regression and CNN models on discharge diagnosis classification. From the table, we can see that CNN model consistently outperform baseline models on all measurement metrics.}
  \label{fig:performance-comparison}
\end{figure}

Figure \ref{fig:performance-comparison} shows the evaluation metrics including accuracy, average or weighted average true positive, false positive, false negative, precision, recall and F1. The exact values of these metrics are shown in Table \ref{tab:diagnosis-classification}.

\begin{table} 
 \centering 
 \begin{tabular}{|l|l|l|l|l|l|} 
 \hline 
Metric & SVM & MLP & LR & RF & CNN \\ \hline  
ACC & 93.71$\pm$0.37 & 91.54$\pm$1.69 & 90.74$\pm$0.14 & 93.69$\pm$0.13 & \textbf{96.11}$\pm$0.08 \\ 
TNR & 96.59$\pm$0.20 & 93.44$\pm$2.76 & 95.21$\pm$0.08 & 96.41$\pm$0.07 & \textbf{97.78}$\pm$0.05 \\ 
FPR & 3.41$\pm$0.20 & 4.56$\pm$0.77 & 4.79$\pm$0.08 & 3.59$\pm$0.07 & \textbf{2.22}$\pm$0.05 \\ 
FNR & 21.61$\pm$0.25 & 29.48$\pm$5.78 & \textbf{15.88}$\pm$0.54 & 32.91$\pm$0.49 & 20.92$\pm$0.62 \\ 
Precision & 60.81$\pm$1.58 & 49.02$\pm$9.87 & 40.41$\pm$0.47 & 64.45$\pm$0.58 & \textbf{78.72}$\pm$0.67 \\ 
Recall & 72.39$\pm$2.53 & 50.52$\pm$12.22 & 68.12$\pm$2.40 & 67.09$\pm$0.49 & \textbf{79.08}$\pm$0.62 \\ 
F1 & 60.86$\pm$1.50 & 47.17$\pm$10.94 & 39.48$\pm$0.54 & 65.04$\pm$0.61 & \textbf{78.45}$\pm$0.50 \\ \hline
WACC & 92.59$\pm$0.14 & 88.49$\pm$3.56 & 86.10$\pm$0.23 & 92.22$\pm$0.17 & \textbf{95.29}$\pm$0.08 \\  
WTNR & 96.59$\pm$0.21 & 90.26$\pm$5.46 & 95.80$\pm$0.11 & 95.73$\pm$0.08 & \textbf{97.13}$\pm$0.10 \\ 
WFPR & 3.41$\pm$0.21 & 4.90$\pm$0.64 & 4.20$\pm$0.11 & 4.27$\pm$0.08 & \textbf{2.87}$\pm$0.10 \\ 
WFNR & 23.38$\pm$0.93 & 30.87$\pm$3.34 & 25.66$\pm$0.55 & 31.96$\pm$0.69 & \textbf{19.06}$\pm$0.36 \\ 
WPrecision & 68.56$\pm$1.86 & 57.68$\pm$8.46 & 53.69$\pm$0.70 & 68.43$\pm$0.64 & \textbf{80.55}$\pm$0.42 \\ 
WRecall & 72.21$\pm$1.39 & 53.18$\pm$11.90 & 64.61$\pm$1.14 & 68.04$\pm$0.69 & \textbf{80.94}$\pm$0.36 \\ 
WF1 & 66.08$\pm$1.23 & 53.15$\pm$11.01 & 45.51$\pm$0.79 & 67.77$\pm$0.72 & \textbf{80.48}$\pm$0.41 \\ 
\hline \end{tabular} 
 \caption{Evaluation metrics of convolutional-based text classification (CNN), support vector machine (SVM) and random forest (RF), multilayer perceptron (MLP) and logistic regression (LR) models on diagnosis classification. The evaluation metric ACC, TNR, FPR, FNR, Precision, Recall, F1 mean accuracy, true negative rate, false positive rate, false negative rate, precision, recall and F1 score respectively. They are computed by averaging the corresponding metric of individual classes. The same metric name with "W" attached at the beginning denote the weighted average of that metric. The weights are sample weights in testing dataset. The standard errors computed by 5 set of results of each model using different random splitting are appended after the metric values with a $\pm$ sign. We see that CNN model consistently outperforms baseline models on all measurements.} 
 \label{tab:diagnosis-classification}
 \end{table}

\begin{figure}
\centering 
  \includegraphics[clip, trim=2.65cm 17cm 1cm 3cm, width=1.00\textwidth]{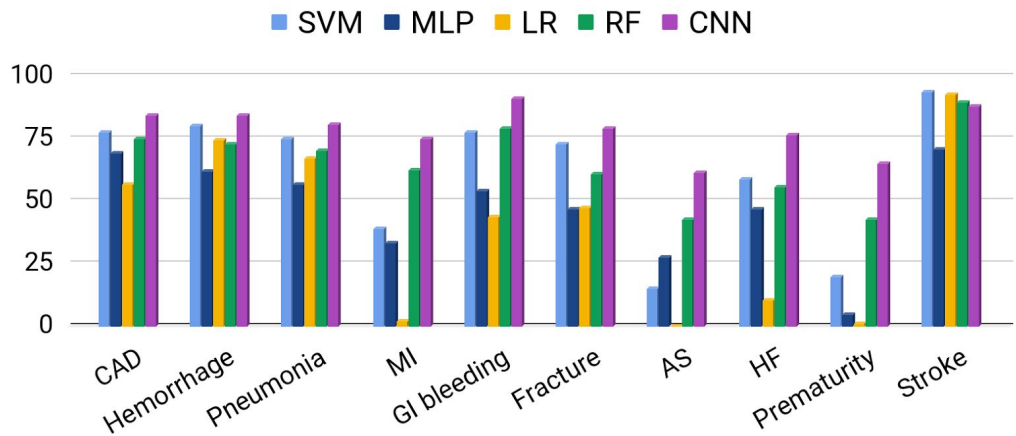}
  \caption{Bar chart demonstrating the F1 score difference on among SVM, random forest, MLP, logistic regression and CNN models on individual discharge diagnosis classification. From the table, we can see that CNN model consistently outperform baseline models on 9 out of 10 diagnosis categories by mostly 10\% to 15\% margins.}
  \label{fig:F1score-comparison}
\end{figure}

\begin{table} 
 \tiny 
 \centering 
 \begin{tabular}{|l|l|l|l|l|l|l|l|l|l|l|l|} 
 \hline
 Model & Label & ACC & TNR & FPR & FNR & Precision & Recall & F1 \\ \hline 
SVM & 0 & 87.41$\pm$0.46 & 97.00$\pm$0.20 & 3.00$\pm$0.20 & 32.66$\pm$1.02 & 91.44$\pm$0.61 & 67.34$\pm$1.02 & 77.54$\pm$0.71 \\ 
MLP & 0 & 75.56$\pm$12.84 & 74.86$\pm$18.72 & 5.14$\pm$1.35 & 36.31$\pm$9.94 & 83.79$\pm$4.29 & 63.69$\pm$9.94 & 68.96$\pm$7.52 \\ 
LR & 0 & 64.08$\pm$0.66 & \textbf{99.73}$\pm$0.11 & \textbf{0.27}$\pm$0.11 & 60.22$\pm$0.59 & \textbf{99.55}$\pm$0.17 & 39.78$\pm$0.59 & 56.84$\pm$0.60 \\ 
RF & 0 & 87.40$\pm$0.32 & 93.47$\pm$0.19 & 6.53$\pm$0.19 & 29.16$\pm$0.68 & 79.85$\pm$0.66 & 70.84$\pm$0.68 & 75.07$\pm$0.55 \\ 
CNN & 0 & \textbf{92.63}$\pm$0.27 & 94.64$\pm$0.44 & 5.36$\pm$0.44 & \textbf{13.93}$\pm$1.26 & 82.86$\pm$1.41 & \textbf{86.07}$\pm$1.26 & \textbf{84.37}$\pm$0.60 \\ 
\hline 
SVM & 1 & 93.87$\pm$0.24 & \textbf{97.72}$\pm$0.25 & \textbf{2.28}$\pm$0.25 & 25.26$\pm$1.56 & 86.54$\pm$1.71 & 74.74$\pm$1.56 & 80.09$\pm$0.58 \\ 
MLP & 1 & 91.69$\pm$1.43 & 94.90$\pm$2.17 & 5.10$\pm$2.17 & 21.39$\pm$5.93 & 66.27$\pm$16.70 & 58.61$\pm$14.87 & 61.97$\pm$15.54 \\ 
LR & 1 & 91.49$\pm$0.24 & 97.66$\pm$0.15 & 2.34$\pm$0.15 & 34.88$\pm$1.25 & \textbf{86.66}$\pm$0.97 & 65.12$\pm$1.25 & 74.34$\pm$1.04 \\ 
RF & 1 & 92.10$\pm$0.28 & 95.79$\pm$0.16 & 4.21$\pm$0.16 & 28.87$\pm$1.43 & 74.81$\pm$1.35 & 71.13$\pm$1.43 & 72.92$\pm$1.34 \\ 
CNN & 1 & \textbf{95.37}$\pm$0.28 & 97.40$\pm$0.14 & 2.60$\pm$0.14 & \textbf{16.10}$\pm$1.71 & 85.00$\pm$0.76 & \textbf{83.90}$\pm$1.71 & \textbf{84.39}$\pm$0.84 \\ 
\hline 
SVM & 2 & 93.10$\pm$0.82 & 96.60$\pm$0.72 & 3.40$\pm$0.72 & 24.85$\pm$5.84 & 77.20$\pm$5.20 & 75.15$\pm$5.84 & 74.63$\pm$1.74 \\ 
MLP & 2 & 90.87$\pm$1.36 & 94.31$\pm$2.00 & 5.69$\pm$2.00 & 26.03$\pm$7.36 & 60.56$\pm$15.27 & 53.97$\pm$13.92 & 56.90$\pm$14.42 \\ 
LR & 2 & 91.96$\pm$0.20 & 94.57$\pm$0.30 & 5.43$\pm$0.30 & 28.05$\pm$1.56 & 63.38$\pm$1.22 & 71.95$\pm$1.56 & 67.31$\pm$0.65 \\ 
RF & 2 & 91.80$\pm$0.29 & 95.96$\pm$0.24 & 4.04$\pm$0.24 & 33.08$\pm$1.38 & 73.41$\pm$1.95 & 66.92$\pm$1.38 & 69.99$\pm$1.50 \\ 
CNN & 2 & \textbf{94.96}$\pm$0.20 & \textbf{97.43}$\pm$0.30 & \textbf{2.57}$\pm$0.30 & \textbf{20.83}$\pm$1.64 & \textbf{82.13}$\pm$2.39 & \textbf{79.17}$\pm$1.64 & \textbf{80.43}$\pm$0.38 \\ 
\hline 
SVM & 3 & 92.61$\pm$0.56 & 93.12$\pm$0.84 & 6.88$\pm$0.84 & 16.33$\pm$4.43 & 28.36$\pm$8.11 & 83.67$\pm$4.43 & 38.90$\pm$8.09 \\ 
MLP & 3 & 89.76$\pm$0.71 & 93.50$\pm$1.12 & 6.50$\pm$1.12 & 46.26$\pm$12.21 & 34.43$\pm$11.69 & 33.74$\pm$9.29 & 33.17$\pm$10.27 \\ 
LR & 3 & 90.81$\pm$0.22 & 90.81$\pm$0.22 & 9.19$\pm$0.22 & \textbf{4.00}$\pm$4.00 & 0.92$\pm$0.25 & \textbf{96.00}$\pm$4.00 & 1.81$\pm$0.49 \\ 
RF & 3 & 93.14$\pm$0.24 & 96.03$\pm$0.27 & 3.97$\pm$0.27 & 36.51$\pm$1.39 & 61.03$\pm$2.22 & 63.49$\pm$1.39 & 62.15$\pm$1.47 \\ 
CNN & 3 & \textbf{95.16}$\pm$0.27 & \textbf{97.18}$\pm$0.22 & \textbf{2.82}$\pm$0.22 & 23.35$\pm$3.13 & \textbf{73.68}$\pm$2.40 & 76.65$\pm$3.13 & \textbf{74.79}$\pm$1.11 \\ 
\hline 
SVM & 4 & 95.28$\pm$1.88 & 98.42$\pm$0.61 & 1.58$\pm$0.61 & 20.89$\pm$10.60 & 83.34$\pm$6.79 & 79.11$\pm$10.60 & 77.68$\pm$5.86 \\ 
MLP & 4 & 93.12$\pm$0.93 & 96.50$\pm$1.31 & 3.50$\pm$1.31 & 31.47$\pm$8.98 & 61.83$\pm$15.86 & 48.53$\pm$12.88 & 54.32$\pm$14.18 \\ 
LR & 4 & 93.60$\pm$0.20 & 93.55$\pm$0.20 & 6.45$\pm$0.20 & \textbf{4.41}$\pm$0.72 & 28.50$\pm$1.02 & \textbf{95.59}$\pm$0.72 & 43.87$\pm$1.24 \\ 
RF & 4 & 96.31$\pm$0.16 & 97.96$\pm$0.14 & 2.04$\pm$0.14 & 20.92$\pm$1.19 & 78.82$\pm$1.58 & 79.08$\pm$1.19 & 78.92$\pm$1.16 \\ 
CNN & 4 & \textbf{98.34}$\pm$0.09 & \textbf{99.09}$\pm$0.13 & \textbf{0.91}$\pm$0.13 & 9.02$\pm$1.56 & \textbf{91.06}$\pm$1.20 & 90.98$\pm$1.56 & \textbf{90.95}$\pm$0.47 \\ 
\hline 
SVM & 5 & 96.14$\pm$0.22 & 97.35$\pm$0.31 & 2.65$\pm$0.31 & 20.33$\pm$3.08 & 67.89$\pm$4.18 & 79.67$\pm$3.08 & 72.88$\pm$2.46 \\ 
MLP & 5 & 93.88$\pm$0.97 & 95.47$\pm$0.96 & 4.53$\pm$0.96 & 28.15$\pm$9.10 & 43.34$\pm$13.30 & 51.85$\pm$14.19 & 46.93$\pm$13.49 \\ 
LR & 5 & 94.31$\pm$0.26 & 94.60$\pm$0.31 & 5.40$\pm$0.31 & \textbf{15.11}$\pm$2.45 & 32.81$\pm$1.38 & \textbf{84.89}$\pm$2.45 & 47.24$\pm$1.45 \\ 
RF & 5 & 94.54$\pm$0.11 & 96.29$\pm$0.14 & 3.71$\pm$0.14 & 31.71$\pm$2.48 & 55.16$\pm$1.02 & 68.29$\pm$2.48 & 60.98$\pm$1.52 \\ 
CNN & 5 & \textbf{96.60}$\pm$0.16 & \textbf{98.26}$\pm$0.30 & \textbf{1.74}$\pm$0.30 & 21.74$\pm$1.02 & \textbf{79.88}$\pm$3.83 & 78.26$\pm$1.02 & \textbf{78.85}$\pm$1.86 \\ 
\hline 
SVM & 6 & 93.02$\pm$0.29 & 93.31$\pm$0.45 & 6.69$\pm$0.45 & 12.15$\pm$7.53 & 10.38$\pm$6.45 & 27.85$\pm$17.09 & 15.02$\pm$9.25 \\ 
MLP & 6 & 91.72$\pm$0.40 & 94.18$\pm$0.42 & 5.82$\pm$0.42 & 47.28$\pm$11.88 & 24.43$\pm$7.14 & 32.72$\pm$8.27 & 27.27$\pm$7.18 \\ 
LR & 6 & 92.60$\pm$0.18 & 92.60$\pm$0.18 & 7.40$\pm$0.18 & \textbf{0.00}$\pm$0.00 & 0.00$\pm$0.00 & 0.00$\pm$0.00 & 0.00$\pm$0.00 \\ 
RF & 6 & 91.65$\pm$0.09 & 95.40$\pm$0.12 & 4.60$\pm$0.12 & 56.62$\pm$0.99 & 42.28$\pm$0.72 & 43.38$\pm$0.99 & 42.80$\pm$0.73 \\ 
CNN & 6 & \textbf{94.50}$\pm$0.38 & \textbf{97.32}$\pm$0.16 & \textbf{2.68}$\pm$0.16 & 40.41$\pm$2.32 & \textbf{63.55}$\pm$2.63 & \textbf{59.59}$\pm$2.32 & \textbf{61.29}$\pm$1.77 \\ 
\hline 
SVM & 7 & 95.29$\pm$0.11 & 95.77$\pm$0.07 & 4.23$\pm$0.07 & 16.10$\pm$2.12 & 44.92$\pm$0.91 & 83.90$\pm$2.12 & 58.49$\pm$1.14 \\ 
MLP & 7 & 94.49$\pm$0.63 & 95.59$\pm$0.93 & 4.41$\pm$0.93 & 23.75$\pm$6.90 & 41.76$\pm$13.25 & 56.25$\pm$14.50 & 46.82$\pm$13.46 \\ 
LR & 7 & 92.96$\pm$0.13 & 92.98$\pm$0.13 & 7.02$\pm$0.13 & \textbf{11.34}$\pm$3.33 & 5.38$\pm$0.54 & \textbf{88.66}$\pm$3.33 & 10.12$\pm$0.95 \\ 
RF & 7 & 94.42$\pm$0.19 & 95.90$\pm$0.19 & 4.10$\pm$0.19 & 32.83$\pm$1.93 & 47.27$\pm$2.70 & 67.17$\pm$1.93 & 55.41$\pm$2.48 \\ 
CNN & 7 & \textbf{96.95}$\pm$0.16 & \textbf{98.15}$\pm$0.09 & \textbf{1.85}$\pm$0.09 & 20.91$\pm$1.47 & \textbf{74.07}$\pm$1.54 & 79.09$\pm$1.47 & \textbf{76.50}$\pm$1.51 \\ 
\hline 
SVM & 8 & 90.96$\pm$3.98 & 97.04$\pm$0.33 & 2.96$\pm$0.33 & 44.30$\pm$18.58 & 27.64$\pm$9.79 & 55.70$\pm$18.58 & 19.85$\pm$5.18 \\ 
MLP & 8 & 95.99$\pm$0.12 & 96.27$\pm$0.19 & 3.73$\pm$0.19 & 25.59$\pm$16.73 & 2.95$\pm$1.65 & 34.41$\pm$19.19 & 4.67$\pm$2.47 \\ 
LR & 8 & 96.19$\pm$0.16 & 96.19$\pm$0.16 & 3.81$\pm$0.16 & \textbf{0.00}$\pm$0.00 & 0.41$\pm$0.25 & 40.00$\pm$24.49 & 0.81$\pm$0.50 \\ 
RF & 8 & 96.54$\pm$0.15 & 97.41$\pm$0.13 & 2.59$\pm$0.13 & 41.53$\pm$2.39 & 33.95$\pm$1.20 & 58.47$\pm$2.39 & 42.85$\pm$1.15 \\ 
CNN & 8 & \textbf{97.76}$\pm$0.23 & \textbf{98.34}$\pm$0.12 & \textbf{1.66}$\pm$0.12 & 22.20$\pm$5.28 & \textbf{56.33}$\pm$3.00 & \textbf{77.80}$\pm$5.28 & \textbf{65.04}$\pm$3.42 \\ 
\hline 
SVM & 9 & \textbf{99.44}$\pm$0.04 & 99.55$\pm$0.03 & 0.45$\pm$0.03 & 3.18$\pm$0.59 & 90.40$\pm$0.72 & 96.82$\pm$0.59 & \textbf{93.49}$\pm$0.42 \\ 
MLP & 9 & 98.29$\pm$0.68 & 98.78$\pm$0.68 & 1.22$\pm$0.68 & 8.52$\pm$6.47 & 70.87$\pm$17.79 & 71.48$\pm$18.88 & 70.62$\pm$17.96 \\ 
LR & 9 & 99.37$\pm$0.05 & 99.37$\pm$0.04 & 0.63$\pm$0.04 & \textbf{0.75}$\pm$0.36 & 86.48$\pm$0.95 & \textbf{99.25}$\pm$0.36 & 92.42$\pm$0.61 \\ 
RF & 9 & 98.96$\pm$0.10 & 99.90$\pm$0.05 & 0.10$\pm$0.05 & 17.81$\pm$1.46 & 97.87$\pm$1.09 & 82.19$\pm$1.46 & 89.33$\pm$1.13 \\ 
CNN & 9 & 98.83$\pm$0.06 & \textbf{99.94}$\pm$0.03 & \textbf{0.06}$\pm$0.03 & 20.72$\pm$1.34 & \textbf{98.63}$\pm$0.60 & 79.28$\pm$1.34 & 87.88$\pm$0.88 \\ 
\hline 
\end{tabular} 
\caption{Evaluation metrics on individual disease classification. The ten class labels correspond to coronary artery disease, hemorrhage, pneumonia, myocardial infarction, gastrointestinal bleeding, fracture, aortic stenos, cardiac failure, prematurity and stroke. The evaluation metrics ACC, TNR, FPR, FNR refer to accuracy, true positive rate, false positive rate, false negative rate. The mapping between label indices and diseases are: 0-coronary artery disease, 1-hemorrhage, 2-pneumonia, 3-myocardial infarction, 4-gastrointestinal bleeding, 5-fracture, 6-aortic stenosis, 7-cardiac failure, 8-prematurity, 9-stroke. The standard errors computed by 5 set of results of each model using different random splitting are appended after the metric values with a $\pm$ sign.}
  \label{tab:individual-diagnosis-classification}
 \end{table}

\subsection{Results on individual diagnosis classification}

The measurement results on individual diagnosis categories are also evaluated and shown in Table \ref{tab:individual-diagnosis-classification}. A bar chart comparing F1 score among the three models is also provided in Figure \ref{fig:F1score-comparison}.

\subsection{T-test on performance across models}
In addition, a t-test has also been done between the weighted average F1 score of the CNN and each of the baseline models, assuming the variances are not equal. The p-values of weighted average F1 metric value of CNN model with that of SVM, random forest, MLP and logistic regression are 1.20e-4, 3.05e-06, 6.81e-2, 1.81e-08 respectively. These values are much less than 0.5, indicating significant difference between the respective performance metrics.
 
\subsection{CNN filter visualization} 
To visualize the patterns learned by each filter category in the convolutional layer of the proposed CNN model, we ranked all phrases in the testing dataset which are scanned by the filter window of convolutional layer by their activation scores in descending order. The top 10 phrases of randomly selected 2 filters per filter size from 3 to 5 is shown in Table \ref{tab:ranked_grams}.

\begin{table}
 \centering 
 \begin{tabular}{l|l}
 \hline
 filter 1 of trigram & filter 2 of trigram\\ \hline
 aortic stenosis dr & hyperlipidemia degenerative joint\\
 aortic stenosis noted & hyperlipidemia obesity tobacco\\
 aortic stenosis referred & hyperlipidemia s p\\
 aortic stenosis bicuspid & hyperlipidemia not known\\
 aortic stenosis hepatitis & hyperlipidemia complete heart\\
 aortic stenosis valve & hyperlipidemia percutaneous coronary\\
 aortic stenosis followed & hyperlipidemia niddm tobacco\\
 aortic stenosis most & hyperlipidemia obesity diabetes\\
 aortic stenosis treated & hyperlipidemia aspirin allergy\\
 aortic stenosis coronary & hyperlipidemia hypertension tobacco\\ 
\hline 
 filter 1 of 4-gram & filter 2 of 4-gram\\
 \hline
 cesarean section delivery membranes & impending respiratory failure injuries\\
spontaneous vaginal delivery in & developed respiratory failure requiring\\
prior to delivery besides & hypoxic respiratory failure successfully\\
by vaginal delivery one & hypoxic respiratory failure discharged\\
term vaginal delivery surgically & use respiratory failure non\\
spontaneous vaginal delivery required & copd respiratory failure requiring\\
section at delivery infant & mother respiratory failure hepatitis\\
induced vaginal delivery apgar & hypoxic respiratory failure a\\
spontaneous vaginal delivery mother & pancytopenia respiratory failure s\\
delivery vaginal delivery apgars & hypoxic respiratory failure transferred\\
 \hline
 filter 1 of 5-gram & filter 2 of 5-gram\\ \hline
 was stable without chest pain & multiple loose watery bm stool\\
catheterization after developing chest pain & small amount of loose stool\\
to have substernal chest pain & noticed red blood around stool\\
flow she had chest pain & her last semi formed stool\\
patient began having chest pain & guaiac positive with red stool\\
her bms no chest pain & have black guaiac positive stool\\
who complained of chest pain & had a well formed stool\\
and developed sharp chest pain & and passed dark black stool\\
onset heavy substernal chest pain & for presumed infectious colitis stool\\
st elevation improved chest pain & dark brown well formed stool\\
\hline \end{tabular}
 \caption{Top 10 3-grams, 4-grams, and 5-grams ranked by activation scores by convolutional filters in the proposed CNN model in descending order. 2 filters per filter size from 3 to 5 are selected. }
 \label{tab:ranked_grams}
 \end{table}

\section{Discussion} 
\subsection{Overall results}

The CNN model we presented achieves, to the best of our knowledge, state-of-the-art prediction performance in discovering complex patterns in unstructured clinical notes for diagnosis decision making. Of the models evaluated, the best performance was achieved with the proposed CNN model that uses pretrained word embeddings and disease coreference resolution. The proposed model achieved 96.11\% accuracy and 80.48\% weighted average F1 score which is around 13\% higher than the best result from baseline models. The proposed model also significantly outperformed traditional machine learning models that rely on bag-of-word features and the PCA feature dimensionality reduction technique. The result suggests the CNN model is suitable for supporting diagnosis decision making in the presence of complex, noisy and unstructured clinical data while at the same time using less layers and parameters that other traditional Deep Network models.

In addition, CNN model achieved close to 80\% precision and recall, while all baseline models have less than 73\% values and are struggling with doing equally well on both measurements. Having relatively equally good performance in precision and recall is important in medical supporting systems since not only the majority diseases should be detected early and accurately, but also the rare diseases need to be identified timely and with confidence. Sometimes, it is even more important to detect rare diseases since it is less likely to detect them and a wrong treatment plan could put the patients' health or even life at risk.

Besides, it is interesting that the Logistic Regression model achieved the minimum false negative rate, something that came at the expense of having the lowest precision value among all models. Intuitively, the model tends to predict a sample with majority category if it is not very confident about whether the sample belongs to a rare class or not, increasing the likelihood of predicting correctly samples labeled with a rare disease class.

\subsection{Individual results}
According to table \ref{tab:individual-diagnosis-classification}, the CNN model achieved the highest F1 score on all 10 disease classes. Furthermore, although some models have relatively high F1 score in categories with large sample size (e.g., coronary artery disease, hemorrhage, pneumonia), they performed badly on categories with less samples (e.g., myocardial infarction, aortic stenosis, prematurity). For example, the SVM achieved an F1 score value of 80.09\%  for the hemorrhage class, only around 4\% lower than the F1 score value of the CNN model, it was only able to achieve F1 score values of 15.02\% on aortic stenosis and 19.85\% on prematurity. On the contrary, the CNN model performed equally well on all disease classes independently of frequently they appear and are diagnosed in the MIMIC corpus.

Table \ref{tab:individual-diagnosis-classification} also indicate that aortic stenosis and prematurity are the hardest disease classes to classify. The highest F1 score values achieved from any of the baseline models for aortic stenosis and prematurity are 42.80\% and  42.85\% respectively. Among all the baseline models, the logistic regression and the MLP model achieved less than 5\% F1 score for the prematurity class. On the other side, the CNN model still achieved F1 score values of 61.29\% and 65.04\% on the same diseases. This fact further confirms that the proposed CNN model works remarkably well at classifying rare (not appearing frequently in the corpus) diseases.
 
The Logistic Regression model had the worst performance among all models with respect to most evaluation metrics. Being essentially a linear model, Logistic Regression usually works well with data that can be separated by a hyperplane. 
However, the latent features of the clinical notes are complex and are very likely not linearly separable. This explains the poor performance of the Logistic Regression model on the discharge diagnosis prediction problem.
  
\subsection{Understanding the model features}
 Last but not least, Table \ref{tab:ranked_grams} shows the top 10 phrases ranked by activation scores by convolutional filters in CNN model. We can see from the table that each filter tends to detect a specific pattern. For example, the first $H_1 \times E= 3 \times 128$ filter detects phrases containing the words ``aortic stenosis", while the second $H_1 \times E= 3 \times 128$ filter detects phrases highly related to "hyperlipidemia". While 3-grams ranked with the highest activation scores by convolutional filters  tend to preserve certain words or phrases, 4-grams and 5-grams assigned with high activation scores, despite very centered around a specific symptom or medical event with different descriptions, are more flexible. For example, the first $H_3 \times E = 5 \times 100$ filter of 5-gram describes chest pain with various conditions such as substernal chest pain, sharp chest pain, onset heavy substernal chest pain and st elevation improved chest pain. We conclude that the convolutional filters or our CNN model are indeed extracting critical patterns for diagnosis, such as symptoms, lab tests, diseases, procedures and abnormalities. It is quite impressive how the CNN model learns to mimic the human clinician's procedure diagnosis decision making.

\section{Future work}

A first direction of future work is to train a CNN model that can distinguish among more than 10 diseases.
The fact that in this work our CNN model uses only 10 classes (corresponding to the 10 most frequently mentioned diseases in the MIMIC corpus) does by no means imply that it is not generalizable. We still have to take care of solving the problem of some diseases appearing and being diagnosed significantly more than others when training our model with more than 10 classes. Towards this end, we can use a slightly different loss function, where the parts of the loss pertaining to different diseases are weighted inversely proportional to the frequency of the respective diseases (penalizing mis-predictions of rare diseases more than mis-predictions of common diseases) and the total loss is a sum of these weighted losses.

A second direction of future work is to extend our CNN model so that it performs \textit{multi-label} classification, predicting more than one diseases per clinical note or \textit{multi-task} classification, predicting, in addition to the disease itself, significant factors such as mortality possibility or severity level of the disease. 

A third direction of future work would be to extend our CNN model so that it supports hierarchical disease prediction. A model that in addition to predicting pneumonia can specify the specific type of pneumonia such as aspiration pneumonia, bacteria pneumonia, hospital-acquired pneumonia, or community-acquired pneumonia for example, could provide a much more useful tool for diagnosis decision making. The performance and success of such a model that supports hierarchical disease prediction highly depends on training on a large corpus that contains adequate samples per disease type and subtype.

A fourth direction of future work is to train our model on a much larger clinical notes corpus, hoping that the benefits of using pretrained word embeddings for the initialization of the CNN model will become more evident. The word embeddings used by our model were trained on the MIMIC dataset that contains only 50000 documents, whereas state-of-the-art word-embedding models are usually trained on millions and billions of documents.


\section{Conclusion}

We have presented a novel data-driven technique for diagnosis prediction, that exploits convolutional neural networks' ability to learn latent features. Unlike many existing works, such as ~\citep{bond2012differential}, ~\citep{grady2011good}, ~\citep{ebell2010ahrq}, and ~\citep{achour2001umls} we did not use any human designed rules (based on prior medical knowledge) for clinical notes' feature learning.
Our approach is flexible in the sense that it can be easily adapted to other datasets or employed in various clinical settings where data availability, characteristics, format and statistical distribution of text vary. Moreover, the fact that it is only based on a subset of the information that is available upon admission time, allows our method to integrate well with the clinical setting workflow and provide timely feedback to the clinician. The efficiency is further improved by the fact that our models can be trained end-to-end, without specific need for fine-tuning hyperparameters of individual components.


\bibliography{ref}

\appendix


\end{document}